\documentclass[10pt,twocolumn,letterpaper]{article}

\usepackage{cvpr}              

\usepackage[accsupp]{axessibility} 
\usepackage{graphicx}
\usepackage{subcaption} 

\definecolor{cvprblue}{rgb}{0.21,0.49,0.74}
\usepackage[pagebackref,breaklinks,colorlinks,allcolors=cvprblue]{hyperref}

\usepackage{multirow}
\usepackage[table]{xcolor} 
\usepackage{pifont}   
\def\paperID{3078} 
\def\confName{CVPR}
\def\confYear{2026}

\title{\vspace{-0.5em}TRCoRSurg: Temporal-Relational Co-Reasoning for Surgical Video Triplet Recognition}

\author{Fang Li$^{1}$\footnotemark[1] \quad Shihao Zou$^{2}$\footnotemark[1] \quad Weixin Si$^{3}$\footnotemark[2] \quad Yang Gao$^{1}$\footnotemark[2] \quad Shuai Li$^{1}$ \quad Aimin Hao$^{1}$ \\
$^{1}$ State Key Laboratory of Virtual Reality Technology and Systems, Beihang University\\
$^2$ Shenzhen Institutes of Advanced Technology, Chinese Academy of Sciences\\
$^3$ Faculty of Computer Science and Control Engineering, Shenzhen University of Advanced Technology \\
\vspace{-1.5em}
}

\definecolor{myblue}{RGB}{30,90,255}

\begin{document}

\maketitle

\renewcommand{\thefootnote}{\fnsymbol{footnote}}
\footnotetext[1]{Co-first authors}
\footnotetext[2]{Corresponding authors}
\renewcommand{\thefootnote}{\arabic{footnote}}

\begin{abstract}

Understanding complex surgical scenes requires recognizing multiple interdependent entities—such as instruments, actions, and targets—and maintaining their relational consistency across time. Existing surgical triplet recognition methods struggle to jointly model intra-frame label dependencies and inter-frame temporal semantics in a unified manner. To address these limitations, we propose a unified framework that integrates spatial, relational, and temporal cues for robust surgical triplet recognition. Specifically, class-specific spatial priors are first extracted through a multi-scale encoder. Then, these priors are refined by a Label Correlation Modeling module with multi-scale class activation map-guided relational extraction (MS-CAMRE), enabling the model to capture both static co-occurrence and dynamic contextual dependencies among triplet components. Furthermore, a Bidirectional Temporal–Relational Fusion Attention (BTRFA) module harmonizes temporal and relational representations to achieve coherent temporal reasoning. We also introduce a new evaluation metric, the Triplet Consistency Error Rate (TCER), which quantitatively measures the model’s capability to preserve causal and semantic consistency across triplets. Extensive experiments on the CholecT45 and ProstaTD datasets show that our method achieves state-of-the-art (SOTA) performance, improving $AP_{IVT}$ by 5.1\% and 7.8\%, respectively. Moreover, on the TCER metric, our approach yields over 36\% and 25\% relative reductions on the two datasets, respectively, underscoring the effectiveness of our framework in temporal–relational co-reasoning. Code is available at \url{https://github.com/Neesky/TRCoRSurg}.

\end{abstract}
\vspace{-5mm}
\section{Introduction}
\vspace{-2mm}
\label{sec:intro}

\begin{figure}[htbp]
    \centering
    \includegraphics[width=\columnwidth]{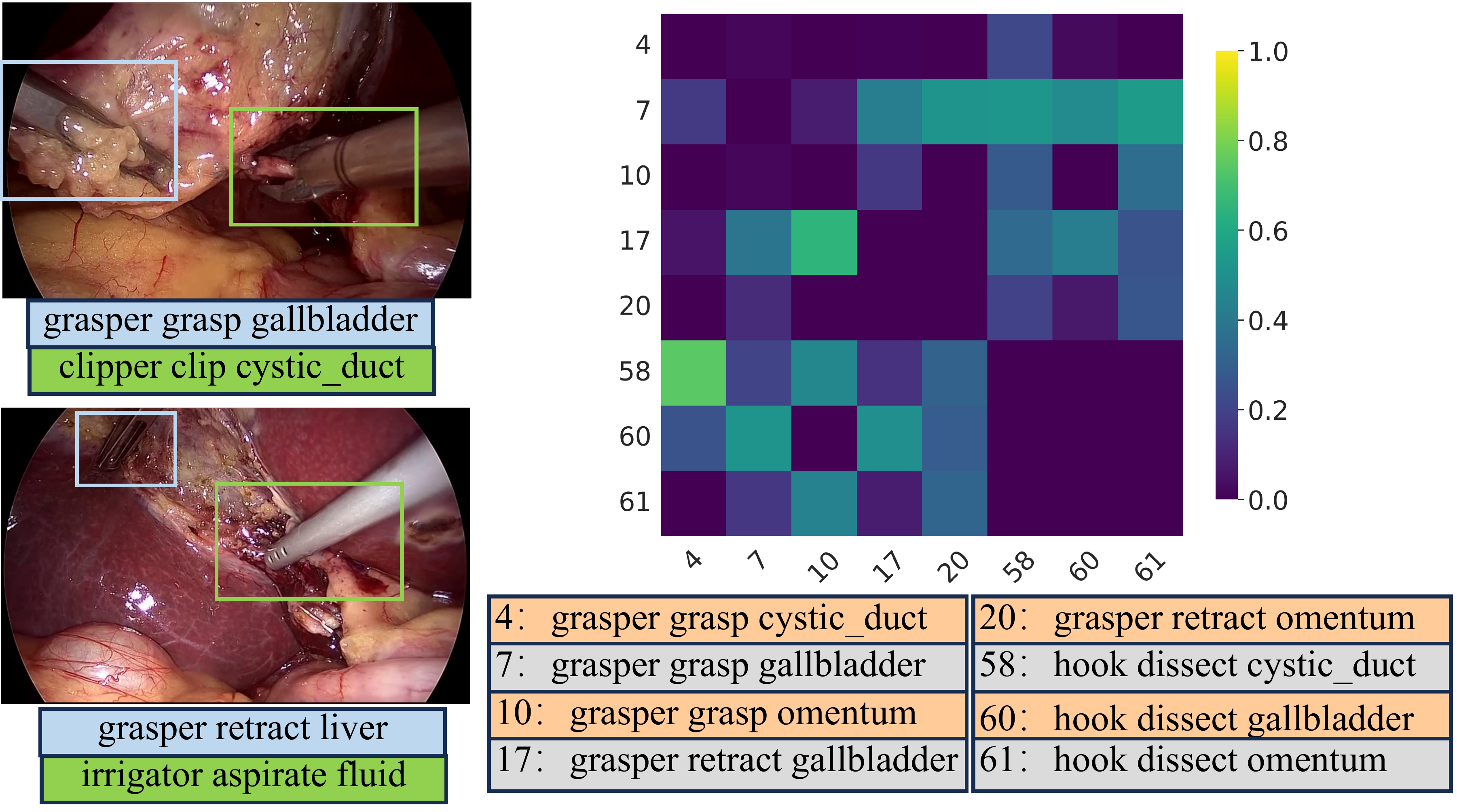}
    \vspace{-6mm}
    \caption{Illustration of label dependencies in surgical triplet recognition. \textbf{Left}: representative frames showing multiple triplets co-occurring within the same scene. \textbf{Right}: pairwise triplet co-occurrence matrices, where the highly uneven distribution indicates strong structured dependencies among triplets.
    }
    \vspace{-5mm}
    \label{fig:relation_heatmap}
\end{figure}

Video-driven fine-grained surgical action recognition aims to identify detailed surgical activities from video frames, providing intra-operative, context-aware assistance that enhances surgical safety~\cite{maier2017surgical}. As a core task in surgical video understanding, it also supports key downstream applications such as data archiving, postoperative analysis, and surgical education~\cite{liu2025deep,zhang2022retrieval,twinanda2016endonet,pei2025instrument,li2024parameter}. Among these fine-grained tasks, surgical triplet recognition has recently emerged as a particularly promising direction, offering the most granular level of activity understanding~\cite{li2024surgical,liu2024surgical,zou2025capturing}. In this paradigm, each surgical activity is represented as a triplet $<$instrument, {verb, target$>$ (IVT), capturing both the entities involved and their underlying semantic relations.

Despite notable advances in surgical video understanding, triplet recognition remains a challenging problem. Existing methods have explored various strategies to improve performance. Attention-based models such as RDV~\cite{nwoye2022rendezvous} learn associative cues from ResNet-derived visual features, while other approaches address long-tailed label distributions~\cite{gui2024tail} or apply knowledge distillation to enhance hierarchical feature learning~\cite{yamlahi2023self,gui2023mt4mtl}. Furthermore, graph-based models~\cite{xi2022forest} and CAM-assisted frameworks~\cite{chen2023surgical} have also been introduced to capture label dependencies and provide coarse spatial priors for IVT triplet prediction.

However, two key limitations remain unresolved: i) \textit{Lack of intra-frame label dependency modeling.} IVT labels exhibit strong anatomical and procedural constraints, forming consistent conditional priors (see \autoref{fig:relation_heatmap}). Although graph-based models~\cite{xi2022forest} capture static label relations, they cannot adapt to the semantic variations that occur within individual frames. Conversely, CAM-based approaches~\cite{chen2023surgical,gui2024tail} provide spatial cues but lack explicit relational reasoning. ii) \textit{Lack of temporal–relational co-reasoning.} Most existing studies treat inter-frame temporal modeling~\cite{gui2023mt4mtl} and intra-frame label dependency learning~\cite{xi2022forest,yamlahi2023self} as independent components. However, simple fusion strategies between them often overlook complementary information and fail to capture the fine-grained spatiotemporal dependencies essential for accurate surgical triplet recognition.

To address these challenges, we propose a unified temporal–relational co-reasoning framework that jointly models intra-frame label dependencies and inter-frame temporal relations for surgical triplet recognition. Specifically, we design a label correlation module that captures both node- and edge-level relationships among instrument, verb, and target labels. At the node level, intrinsic semantic priors (ISP) are fused with variant activation cues (VAC) to integrate stable semantic meaning with variant visual evidence. At the edge level, an MS-CAMRE module adaptively learns edge correlations guided by triplet co-occurrence matrices as priors, ensuring relational stability and contextual adaptability. To explicitly couple temporal and relational reasoning, we further introduce the BTRFA module, which integrates inter-frame temporal memory and intra-frame relational attention for coherent and context-aware decision refinement. Moreover, since existing methods~\cite{yamlahi2023self,xi2023chain} primarily evaluate per-entity or per-triplet accuracy while overlooking the evaluation of combination consistency among IVT entities, we propose the TCER metric to quantify compositional inconsistencies, providing a more comprehensive evaluation of relational reasoning performance.

Our contributions can be summarized as follows:

\begin{itemize}
    \item We introduce a \textit{unified temporal-relational co-reasoning framework} that jointly models intra-frame label dependencies and inter-frame temporal relations through a bidirectional fusion attention mechanism, enabling coherent surgical triplet reasoning across frames. 
    
    \item We propose a \textit{label correlation module} that jointly models node- and edge-level dependencies among triplets, where nodes integrate visual semantic evidence, and edges adaptively model triplet relationships guided by co-occurrence matrices as relational priors.

    \item We introduce a \textit{new TCER metric} to evaluate the compositional consistency of IVT predictions. Extensive experiments on two benchmark datasets demonstrate that our method achieves SOTA performance, yielding over a 5\% relative improvement in triplet recognition accuracy and more than a 25\% relative reduction in TCER.

\end{itemize}

\vspace{-2mm}
\section{Related Work}
\vspace{-2mm}
\label{sec:related_work}

\textbf{Multi-label image classification} aims to predict semantic labels for each instance while capturing their co-occurrence and inter-label dependencies. Graph-based frameworks have shown particular effectiveness in this task. \citet{DBLP:conf/cvpr/ChenWWG19} used GCNs to learn structured label representations, and Wang et al.~\cite{DBLP:journals/corr/abs-2008-08407} combined GloVe embeddings with co-occurrence priors for adaptive label-aware learning. Object-GCN~\cite{zhang2024hierarchical} further integrated object detection with GCNs to jointly capture semantic and visual correlations, while Zhu et al.~\cite{DBLP:conf/iccv/ZhuLLGLC23} highlighted the scene-dependent nature of label co-occurrence. These graph-driven and knowledge-guided methods are well-suited for triplet recognition, as they effectively model the contextual co-occurrence and label dependencies among triplet entities.

\textbf{Action recognition} aims to identify human- or tool-driven activities in videos by modeling spatiotemporal dependencies. Early 3D CNN-based frameworks~\cite{wang2016temporal, carreira2017quo} jointly captured motion and appearance cues, later extended with self-attention and Transformer architectures for long-range temporal modeling~\cite{wang2018non}. MViTv2~\cite{li2022mvitv2} introduced hierarchical spatiotemporal representations, while UniformerV2~\cite{li2023uniformerv2} achieved efficient dynamic token aggregation for fine-grained motion modeling. More recent methods integrate multi-modal cues such as RGB, optical flow, and depth~\cite{wang2024multifuser, lai2024smart} to improve temporal alignment and semantic consistency. These advances directly inspire surgical video understanding, where precise modeling of instrument motion and procedural context is crucial.

\textbf{Surgical triplet recognition} aims to jointly predict $<$\textit{instrument}, \textit{verb}, \textit{target}$>$ triplets, providing a structured understanding of intra-operative activities. Rendezvous~\cite{nwoye2022rendezvous} introduced an instrument-centric formulation that uses CAMs to condition verb and target prediction, while its temporal extension, RIT~\cite{sharma2023rendezvous}, leveraged inter-frame dependencies for improved verb recognition. Chen et al.~\cite{chen2023surgical} further disentangled sub-tasks through CAM-guided attention and self-distillation to reduce cross-instrument interference. Recent advances have shifted toward relational reasoning and generative modeling: Liu et al.~\cite{liu2024surgical} proposed a diffusion-based framework for coherent triplet prediction, Xi et al.~\cite{xi2022forest, xi2023chain} employed GCNs and biomedical language models to enhance semantic consistency, and Gui et al.~\cite{gui2023mt4mtl, gui2024tail} introduced instance-level disentanglement and contrastive learning to mitigate class imbalance. 
However, these methods often treat temporal and relational cues in isolation, lacking a unified framework to jointly model spatiotemporal dependencies for comprehensive surgical video triplet recognition.

\vspace{-1mm}
\section{Method}
\vspace{-2mm}
\label{sec:method}

\begin{figure*}[t]
    \centering
    \includegraphics[width=\textwidth]{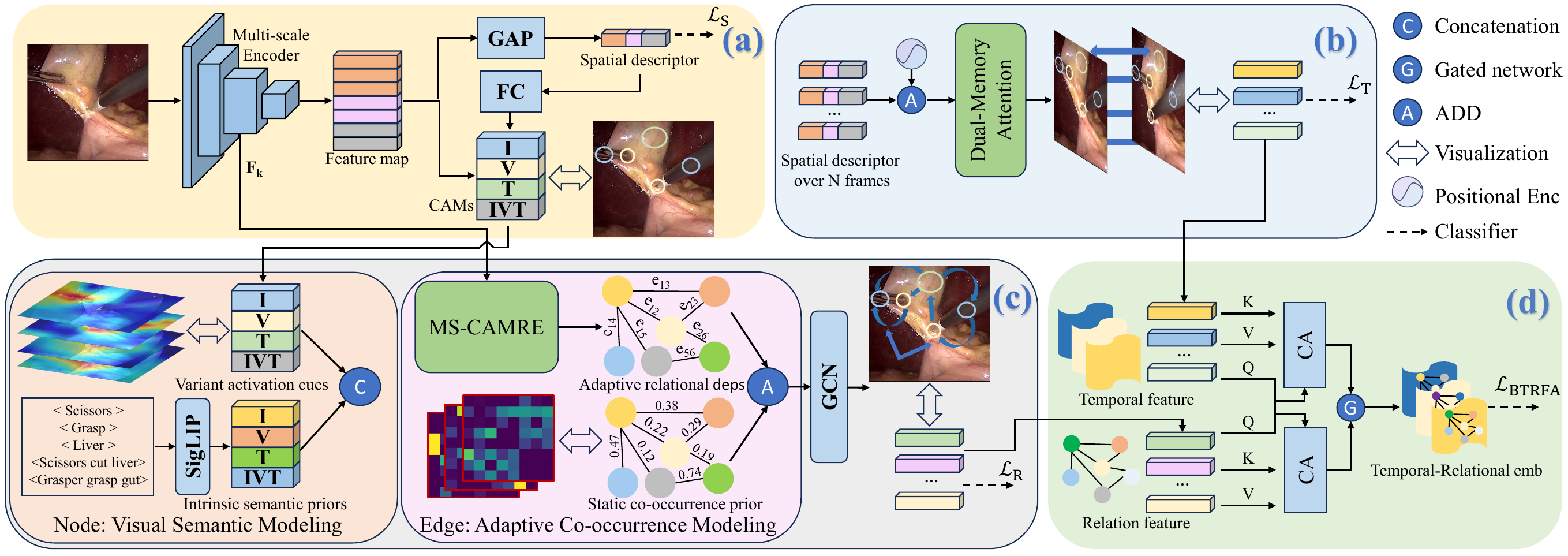}
    \vspace{-7mm}
    \caption{
        Framework Overview. \textbf{(a) Multi-Scale Encoder} extracts multi-scale features, pools them into a global spatial descriptor to capture holistic semantics, and generates CAMs to guide subsequent GCN-based relational reasoning. \textbf{(b) Dual-Memory Temporal Attention} fuses spatial descriptors across $N$ frames for temporal reasoning. \textbf{(c) Label Correlation Modeling} integrates visual semantic evidence at the node level, while the MS-CAMRE module adaptively models triplet relationships at the edge level, guided by co-occurrence matrices as relational priors. \textbf{(d) Temporal–Relational Fusion} employs the BTRFA module to jointly reason over label dependencies and temporal relationships through gated fusion, enabling coherent and context-aware triplet prediction.
    }
    \label{images:overview}
    \vspace{-8pt}
\end{figure*}

Our framework for surgical triplet recognition comprises four key components, as illustrated in~\autoref{images:overview}. i) The Multi-Scale Encoder (Sec.~\ref{sec:spatial-temporal}) extracts hierarchical visual features and pools them into a global spatial descriptor. ii) The Dual-Memory Temporal Attention (Sec.~\ref{sec:spatial-temporal}) fuses spatial descriptors across consecutive frames. iii) Label Correlation Modeling (Sec.~\ref{sec:correlation}) integrates visual semantic evidence at the node level and adaptively models triplet relationships at the edge level. iv) Temporal–Relational Fusion (Sec.~\ref{sec:temporal_relation}) employs the BTRFA module to jointly reason over label dependencies and temporal relationships through gated fusion. Finally, we introduce the TCER metric (Sec.~\ref{sec:TCER}) to quantitatively evaluate the compositional consistency of predicted triplet entities.

\vspace{-1mm}
\subsection{Spatiotemporal Feature Encoding}
\label{sec:spatial-temporal}
\vspace{-1mm}

As illustrated in \autoref{images:overview}~(a), given an input frame $\textbf{x} \in \mathbb{R}^{3 \times H \times W}$, the spatial encoder extracts a set of multi-scale feature maps $\{\mathbf{F}^k \in \mathbb{R}^{D_k \times H_k \times W_k}\}$ from successive backbone stages, where $k \in \{1, 2, 3\}$. For brevity, we denote the highest-level feature map as $\mathbf{F} \triangleq \mathbf{F}^3$. These multi-scale features progressively capture both low-level texture information and high-level semantic representations, which are essential for fine-grained multi-label and triplet-based recognition in surgical videos. A global average pooling (GAP) operation is then applied to $\mathbf{F}$ to produce a compact spatial descriptor $\mathbf{d} = \mathrm{GAP}(\mathbf{F}) \in \mathbb{R}^{D}$, serving as a holistic embedding for multi-label prediction.

To extract class-specific spatial evidence, we compute a CAM for each triplet entity label $c$ as
\begin{equation}
    \mathrm{CAM}_c = \sum_{i=1}^{D} w_{c,i}\,\mathbf{F}_i,
\end{equation}
where $w_{c,i}$ denotes the classifier weight associated with the $i$-th feature channel, indicating its contribution to class $c$. The resulting CAMs are aggregated into a tensor $\mathrm{CAM} \in \mathbb{R}^{C \times D}$, where $C$ denotes the total number of triplet entities. These activation maps provide interpretable spatial priors for subsequent modules by highlighting anatomically relevant regions that support each prediction.

Inspired by SAM2 \cite{ravi2024sam}, we introduce a lightweight dual-memory attention mechanism (see Fig. \ref{images:overview}b) to improve temporal coherence. This module leverages two recurrent buffers: a semantic memory for the last N spatial descriptors and a decision memory for the last N output logits. Through cross-attention, the current descriptor fuses with these memories to produce the temporal feature $\mathbf{F}^T$. By capturing explicit, semantically aware temporal dependencies, our approach maintains stable and coherent IVT triplet predictions, particularly when processing low-frame-rate surgical footage.

\vspace{-1mm}
\subsection{Label Correlation Modeling}
\label{sec:correlation}
\vspace{-1mm}

Triplets in surgical videos exhibit strong spatiotemporal contextual dependencies, as their distributions are governed by anatomical structures and procedural workflows, \ie, specific instruments and actions frequently co-occur to accomplish particular surgical intents. To explicitly exploit these constraints, we introduce label correlation modeling, as illustrated in \autoref{images:overview}~(c), which constructs a graph from two complementary perspectives: nodes for visual semantic modeling and edges for adaptive co-occurrence modeling.

\textbf{Nodes for Visual Semantic Modeling.} Existing methods typically employ static word embeddings (\eg, GloVe) as graph nodes~\cite{xi2022forest}, capturing only the static semantics of each entity. However, in surgical scenes, relational cues among entities vary dynamically with context, making it essential to model such contextual variations in label correlations. To address this, we construct node representations from two aspects: variant activation cues $v$ derived from CAMs, which capture variant visual evidence for each entity label, and intrinsic semantic priors $e$ obtained from text encoder, which preserve the static semantic meaning. 

Formally, for entity label $c$ at frame $t$, we have
\begin{gather}
    \mathbf{e}_c = \text{Norm}\big(\text{Proj}_e \big(\text{SigLIP}_{\text{text}}(\text{Prompt}(c))\big)\big), \\
    \mathbf{v}_{c,t} = \text{Norm}\big(\text{Proj}_v\big(\text{CAM}_{c,t}\big)\big).
\end{gather}
Here, $\text{SigLIP}_{\text{text}}(\cdot)$\cite{zhai2023sigmoid} denotes a pretrained text encoder that extracts semantic priors from the class-specific prompt $\text{Prompt}(c)$, while $\text{CAM}_{c,t}$ represents the class activation map for label $c$ at frame $t$. The linear projections $\text{Proj}_e$ and $\text{Proj}_v$ map textual and visual embeddings into a shared latent space, and $\text{Norm}(\cdot)$ denotes feature normalization. Finally, the node for label $c$ at frame $t$ is constructed by concatenating the semantic prior and visual activation as $\mathbf{F}^{\text{label}}_{c,t} =[\mathbf{e}_c, \mathbf{v}_{c,t}]$. This representation jointly encodes label-level semantics and frame-specific visual evidence, serving as the node embedding for subsequent relational reasoning in the GCN.

\textbf{Edges for Adaptive Co-occurrence Modeling.} To model adaptive relational dependencies, we introduce the MS-CAMRE module, as illustrated in \autoref{images:MS_CAMRE}. This module employs CAMs as queries to guide relation extraction, generating semantically meaningful and interpretable edge features. To further ensure training stability, a zero-convolution mechanism is incorporated to progressively and adaptively inject dynamic edge weights during learning.

Specifically, given multi-scale feature maps ${\mathbf{F}^k}$ and the class-specific $\text{CAM}_c \in \mathbb{R}^D$, we first apply a 2D convolutional projection to align the channel dimensions of all feature maps, ensuring compatibility across scales. Subsequently, multi-level relational information $\{\mathbf{z}_c^{k}\}$ for entity label $c$ is extracted through CAM-guided cross-attention over the multi-scale features:
\begin{equation}
    \mathbf{z}_c^{k} = \text{Attn}\left(Q = \text{CAM}_c,; K = \mathbf{F}^k,; V = \mathbf{F}^k\right).
\end{equation}
To consolidate relational cues across scales, we employ a lightweight squeeze-and-excitation (SE) fusion strategy. Global descriptors pooled from $\{\mathbf{z}_c^{k}\}$ generate channel-wise weights that emphasize informative relations and suppress redundancy. The fused features are then aggregated into a unified label-level representation, with a zero-initialized convolution applied for stable and progressive optimization:
\begin{equation}
    \mathbf{z}_c = \text{Conv}_{\text{zero}}\big(\text{SE-Fuse}(\{\mathbf{z}_c^k\})\big).
\end{equation}

We further incorporate static semantic priors into edge modeling by constructing a dataset-level co-occurrence matrix $\mathbf{M} \in \mathbb{R}^{C \times C}$ using the Jaccard similarity of label occurrences. This matrix encodes empirical co-occurrence dependencies among surgical triplets, serving as a relational prior. The adaptive edge weights $\mathbf{w}$ are then computed as:
\begin{equation}
\mathbf{w}_{ij} = \sigma\left(\mathbf{z}_{c_i}^\top \mathbf{z}_{c_j} + \mathbf{M}_{c_ic_j}\right),
\end{equation}
where $\mathbf{z}_{c_i}^\top \mathbf{z}_{c_j}$ represents the adaptive correlation strength between label $c_i$ and $c_j$, and $\sigma(\cdot)$ denotes the sigmoid normalization function. After going through a GCN, we obtain the relational feature denoted as $\mathbf{F}^R=\text{GCN}(\mathbf{F}^{\text{label}}, \mathbf{w})$.

\begin{figure}[t]
    \centering
    \includegraphics[width=\columnwidth]{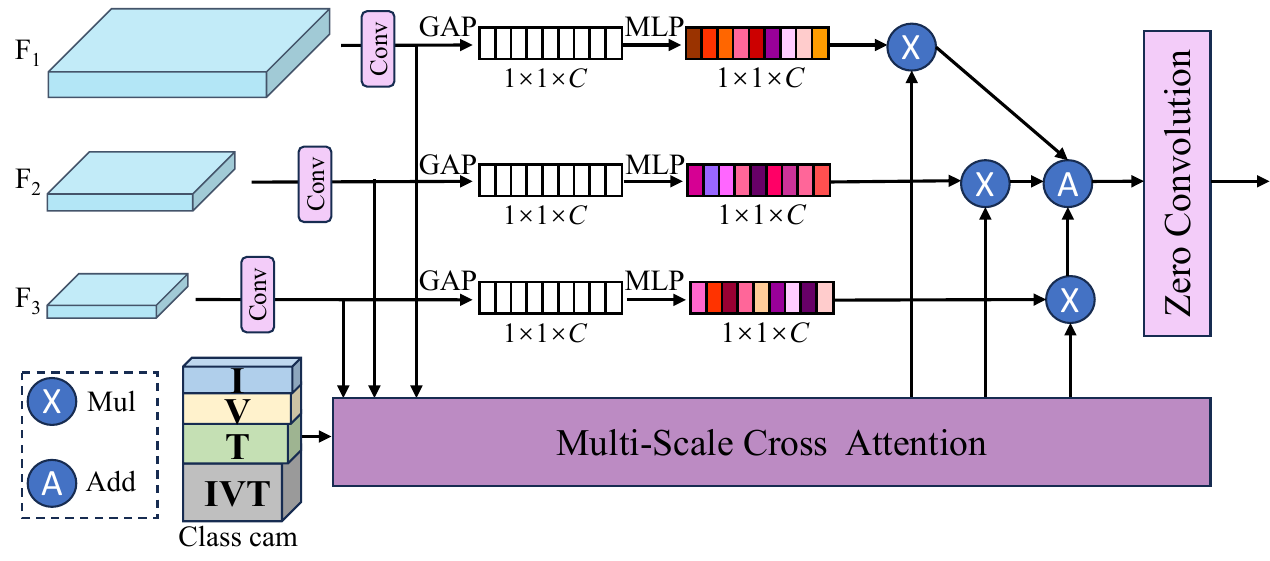}
    \caption{MS-CAMRE Module Overview. The module integrates multi-scale CAMs to guide relational extraction, generating semantically interpretable edges between nodes. A zero-convolution mechanism progressively injects adaptive edge weights during training, ensuring stable and flexible optimization.
    }
    \label{images:MS_CAMRE}
\end{figure}

By integrating dataset-level co-occurrence priors with CAM-guided relational cues, the proposed label correlation modeling enables stable, context-aware reasoning across IVT entities, enhancing the robustness and interpretability of surgical triplet recognition.

\vspace{-1mm}
\subsection{Temporal-Relational Fusion}
\label{sec:temporal_relation}
\vspace{-1mm}

We propose the BTRFA module (\autoref{images:overview}~(d)) to jointly capture temporal context and label correlations for surgical triplet recognition. Given the temporal feature $\mathbf{F}^{T}$ from \autoref{images:overview}~(b) and the relation-aware feature $\mathbf{F}^{R}$ from \autoref{images:overview}~(c), two symmetric cross-attention branches are applied to exchange contextual information:
\begin{gather}
    \hat{\mathbf{F}}^{T} = \mathrm{Attn}(Q=\mathbf{F}^{T}, K=\mathbf{F}^{R}, V=\mathbf{F}^{R}), \\
    \hat{\mathbf{F}}^{R} = \mathrm{Attn}(Q=\mathbf{F}^{R}, K=\mathbf{F}^{T}, V=\mathbf{F}^{T}),
\end{gather}
To adaptively balance their contributions, a learnable gated fusion mechanism is applied, where concatenated features pass through a linear layer followed by a sigmoid activation to generate a gate map:
\begin{gather}
    \mathbf{g} = \sigma\big(\text{FC}([\mathbf{\hat{F}}^T, \mathbf{\hat{F}}^R])\big),\\
    \mathbf{\hat{F}} = \mathbf{g} \odot \mathbf{\hat{F}}^T + (1-\mathbf{g}) \odot \mathbf{\hat{F}}^R,
\end{gather}
where $\sigma(\cdot)$ denotes the sigmoid function and $\odot$ represents element-wise multiplication.

Through this bidirectional interaction, the two streams evolve collaboratively rather than sequentially: the temporal branch captures semantic consistency along the surgical timeline, while the relational branch enforces context-aware regularization among correlated labels. The fused representation thus effectively unifies temporal and relational reasoning, reducing frame-level ambiguity and enhancing the robustness and consistency of surgical triplet recognition.

\vspace{-1mm}
\subsection{Loss Function}
\label{sec:loss}
\vspace{-2mm}

The training process of our framework consists of two stages. In the first stage, the spatial encoder in \autoref{images:overview}~(a) is trained using the loss function
\begin{equation}\mathcal{L}_1 = \mathcal{L}_S,\end{equation}
where $\mathcal{L}_S$ is derived from the spatial descriptor $\mathbf{F}$ through a fully connected classifier.
In the second stage, the remaining three modules are jointly trained using a loss function:
\begin{equation}
\mathcal{L}_{2} = \mathcal{L}_{T} + \mathcal{L}_{\mathrm{R}}+ \mathcal{L}_{\mathrm{BTRFA}},
\end{equation}
where $\mathcal{L}_{T}$, $\mathcal{L}_{\mathrm{R}}$, and $\mathcal{L}_{\mathrm{BTRFA}}$ correspond to the temporal modeling, label correlation, and temporal–relational fusion modules, respectively, as illustrated in \autoref{images:overview}~(b–d). Similar to the first stage, we pass the feature output of each module ($\mathbf{F}^T$, $\mathbf{F}^R$, or $\mathbf{\hat F}$) through an FC classifier to compute the corresponding loss.

Every loss $\mathcal{L}$ can be composed of two primary components: a direct classification loss $\mathcal{L}_{\mathrm{entity}}$ and a coupling loss $\mathcal{L}_{\mathrm{couple}}$, 
formulated as:
\vspace{-2px}
\begin{equation}
\mathcal{L} = \mathcal{L}_{\mathrm{entity}} + \mathcal{L}_{\mathrm{couple}}.
\end{equation}
\vspace{-2px}
For $\mathcal{L}_{\mathrm{entity}}$, we apply binary cross-entropy (BCE) directly over the set of all triplet candidates $\mathcal{C}$:
\vspace{-2px}
\begin{equation}
\mathcal{L}_{\mathrm{entity}} = \sum_{c \in \mathcal{C}} \mathrm{BCE}(\hat{y}_{c}, y_{c}),
\end{equation}
\vspace{-2px}
where $y_{c}$ denotes the ground-truth label for triplet $c$.The coupling loss $\mathcal{L}_{\mathrm{couple}}$ is designed to supervise the atomic components (Instrument $I$, Verb $V$, and Target $T$) using triplet-level predictions. For a specific atomic label $a$ belonging to category $k \in \{I, V, T\}$, we derive its predicted probability by taking the maximum value among all triplets that contain that label:
\vspace{-2px}
\begin{equation}
\hat{P}_{a}^k = \max\limits_{c \in \mathcal{C}, c \ni (k,a)} \hat{y}_{c}.
\end{equation}
\vspace{-2px}
where $c \ni (k,a)$ denotes the set of all triplets $c$ that contain the label $a$ from category $k$.
The coupling loss is then defined as the sum of BCE losses across all atomic labels:
\begin{equation}
\mathcal{L}_{\mathrm{couple}} = \sum_{k \in {I, V, T}} \sum_{a \in \mathcal{A}^k} \mathrm{BCE}(\hat{P}_{a}^k, y_{a}^k),\end{equation}where $\mathcal{A}^k$ is the set of labels for category $k$, and $y_{a}^k$ is the corresponding ground-truth.

\subsection{Triplet Consistency Error Rate}
\label{sec:TCER}

\begin{figure}[h]
    \centering
    \includegraphics[width=\columnwidth]{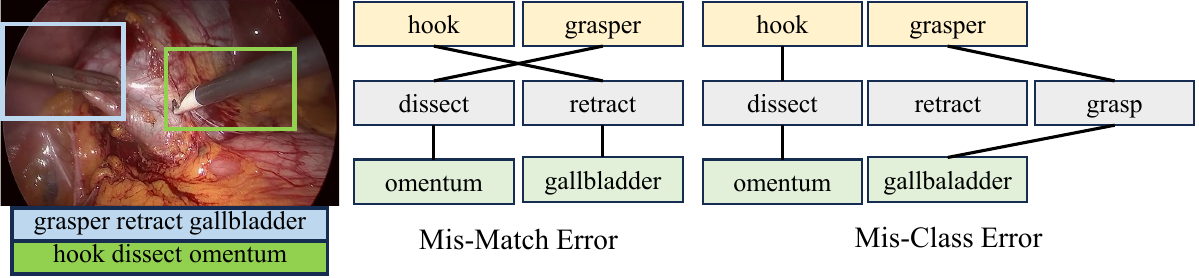}
    \caption{Illustration of proposed TCER metric. The left image shows a video frame, while the right panels depict the Match Error and Direct Error components in TCER, respectively.}
    \label{fig:tcer}
\end{figure}

Conventional metrics typically measure either entity-level or triplet-level accuracy, overlooking cases where individual entities are correctly predicted but their combinations are not. This limitation indicates that a model may recognize entities yet fail to capture their relational dependencies. To explicitly evaluate such reasoning capability, we propose the Triplet Consistency Error Rate (TCER) of two types:
\begin{equation}
    TCER_{\text{m}}=\frac{N_{\text{mis-match}}}{N_{\text{marginal}}}, \quad TCER_{\text{c}}=\frac{N_{\text{mis-class}}}{N_{\text{marginal}}},
\end{equation}
where $N_{\text{marginal}}$ denotes the number of frames where all component predictions are correct. (\eg, $\mathcal{T}_{gt}=\{(I_1,V_1,T_1), (I_2,V_2,T_2)\}$). $N_{\text{mis-match}}$ counts cases in which all entities are correctly identified but incorrectly paired across triplets (\eg, $(I_1,V_2,T_1)$), while $N_{\text{mis-class}}$ measures frames where at least one element in the predicted triplet does not belong to the ground-truth entity set (\eg, $(I_1,V_3,T_1)$). Detailed examples are presented in \autoref{fig:tcer}.

By decomposing relational inconsistencies into these two categories, TCER offers a fine-grained diagnostic measure of inter-entity consistency and reasoning fidelity. This evaluation is especially valuable in medical multi-label classification, where incorrect combinations, such as pairing a correct instrument with an inappropriate verb, can lead to clinically implausible or unsafe interpretations.

\section{Experiments}
\label{sec:Experiments}

\begin{table*}[t]
    \centering
    \scriptsize
    \caption{Quantitative results on CholecT45 and ProstaTD datasets. 
    Higher values indicate better performance for AP and Top-$k$ metrics, while lower values are preferred for TCER$_{m}$ and TCER$_{c}$.}
    \begin{tabular}{c|l|c|ccccccc|cc}
        \toprule
            Dataset & Method & $AP_{IVT}$$\uparrow$ & $AP_{I}$$\uparrow$ & $AP_{V}$$\uparrow$ & $AP_{T}$$\uparrow$ & $Top_5$$\uparrow$ & $Top_{10}$$\uparrow$ & $Top_{20}$$\uparrow$ & & TCER$_{m}$$\downarrow$ & TCER$_{c}$$\downarrow$ \\
        \midrule
        \multirow{7}{*}{CholecT45}
            & TERL~\cite{gui2024tail} & 38.9 ± 2.5 & 93.5 ± 2.4 & 72.8 ± 2.8 & \textbf{51.3 ± 3.8} & -- & -- & -- & & -- & -- \\
            & RIT~\cite{sharma2023rendezvous} & 33.8 ± 2.5 & 91.2 ± 1.9 & 65.3 ± 2.8 & 43.7 ± 1.6 & 81.7 & 90.6 & 95.2 & & 8.71 & 2.80 \\
            & MT4MTLKD~\cite{gui2023mt4mtl} & 37.1 ± 0.5 & 93.1 ± 2.1 & 71.8 ± 3.4 & 48.8 ± 3.8 & 83.2 ± 0.9 & 91.9 ± 0.6 & 94.8 ± 0.3 & & 6.50 & 2.46 \\
            & SDSwin~\cite{yamlahi2023self} & 36.9 ± 2.4 & 92.5 ± 1.7 & 67.3 ± 2.4 & 45.2 ± 1.5 & 83.2 ± 0.8 & 91.1 ± 0.9 & 95.7 ± 0.4 & & 8.85 & 2.24 \\
            & CoLSurgical~\cite{xi2023chain} & 38.2 & 94.1 & 62.5 & 41.9 & 84.5 & 92.4 & \textbf{97.2} & & -- & -- \\
            & RDV~\cite{nwoye2022rendezvous} & 29.9 & 92.0 & 60.7 & 38.3 & 76.3 & 88.7 & 95.9 & & 7.20 & 3.10 \\
            & \textbf{Ours} & \textbf{40.9 ± 2.8} & \textbf{95.7 ± 1.7} & \textbf{73.6 ± 1.8} & 51.1 ± 2.6 & \textbf{86.4 ± 0.7} & \textbf{92.6 ± 0.6} & 96.2 ± 0.5 & & \textbf{4.16} & \textbf{2.09} \\
        \midrule
            \multirow{5}{*}{ProstaTD}
            & RIT~\cite{sharma2023rendezvous} & 28.3 & 75.4 & 57.2 & 51.8 & 59.1 & 77.3 & 85.1 & & 19.50 & 5.20 \\
            & MT4MTLKD~\cite{gui2023mt4mtl} & 35.2 & 84.1 & 65.3 & 60.2 & 64.3 & 80.2 & 87.1 & & 17.70 & 4.90 \\
            & SDSwin~\cite{yamlahi2023self} & 34.8 & 82.4 & 61.3 & 56.8 & 62.2 & 77.9 & 88.4 & & 18.10 & 5.00 \\
            & RDV~\cite{nwoye2022rendezvous} & 26.1 & 74.2 & 55.1 & 49.8 & 57.0 & 75.9 & 84.1 & & 20.80 & 5.60 \\
            & \textbf{Ours} & \textbf{37.5} & \textbf{88.0} & \textbf{65.9} & \textbf{61.7} & \textbf{68.1} & \textbf{81.3} & \textbf{91.6} & & \textbf{13.20} & \textbf{4.70} \\
        \bottomrule
    \end{tabular}
    
    \label{tab:Quantitative_AP_TCER}
\end{table*}


\textbf{Benchmark Datasets.} This paper employs two benchmark datasets for triplet-based surgical action recognition: the CholecT45 dataset\cite{nwoye2023cholectriplet2021} and the ProstaTD dataset \cite{chen2025prostatd}. 
The CholecT45 dataset comprises 45 laparoscopic cholecystectomy video sequences recorded at 1~fps, yielding a total of 100.9K frames and 161K annotated triplet instances. Each frame is labeled with 100 triplets, defined over 6 instruments, 10 verbs, and 15 targets.
The ProstaTD dataset is a large-scale, multi-institutional benchmark designed for robot-assisted radical prostatectomy analysis. It includes 21 surgical videos totaling 60,529 annotated frames and 165,567 structured triplet instances collected from three heterogeneous domains. Each frame is annotated with 89 triplets, composed of 7 instruments, 10 verbs, and 10 targets.

\textbf{Evaluation Metrics.} The model performances are evaluated using average precision (AP) metrics, which are standard in previous surgical triplet recognition works~\cite{nwoye2023cholectriplet2021, chen2025prostatd}. 
The AP metrics are organized into three categories: 
(1) Triplet-level AP (AP$_{\text{IVT}}$) measuring the correctness of complete instrument--verb--target (IVT) triplets; 
(2) Component-level AP ($AP_I$, $AP_V$, $AP_T$) evaluating the precision of individual instruments, verbs, and targets; 
and (3) Top-$K$ accuracy ($Top_5$, $Top_{10}$, $Top_{20}$) capturing whether ground-truth triplets appear among the top-$K$ ranked predictions. 
The main metric is AP$_{\text{IVT}}$, reflecting overall triplet recognition performance.

\textbf{Implementation Details.} All experiments are conducted on four NVIDIA H100 GPUs using the Adam optimizer for network optimization. The training procedure follows a two-stage paradigm. In the first stage, the model is trained with a batch size of 32 and an initial learning rate of $3\times10^{-4}$ for 10 epochs to pretrain the feature extraction modules. In the second stage, fine-tuning is performed with the same batch size of 32 but a reduced learning rate of $2\times10^{-5}$ for 10 epochs to jointly optimize the unified Temporal-Relational Co-Reasoning framework. Following our architecture design, the memory temporal depth $N$ is empirically set to 8.

\begin{figure*}[tb]
    \centering
    \vspace{-2mm}
    \includegraphics[width=0.96\textwidth]{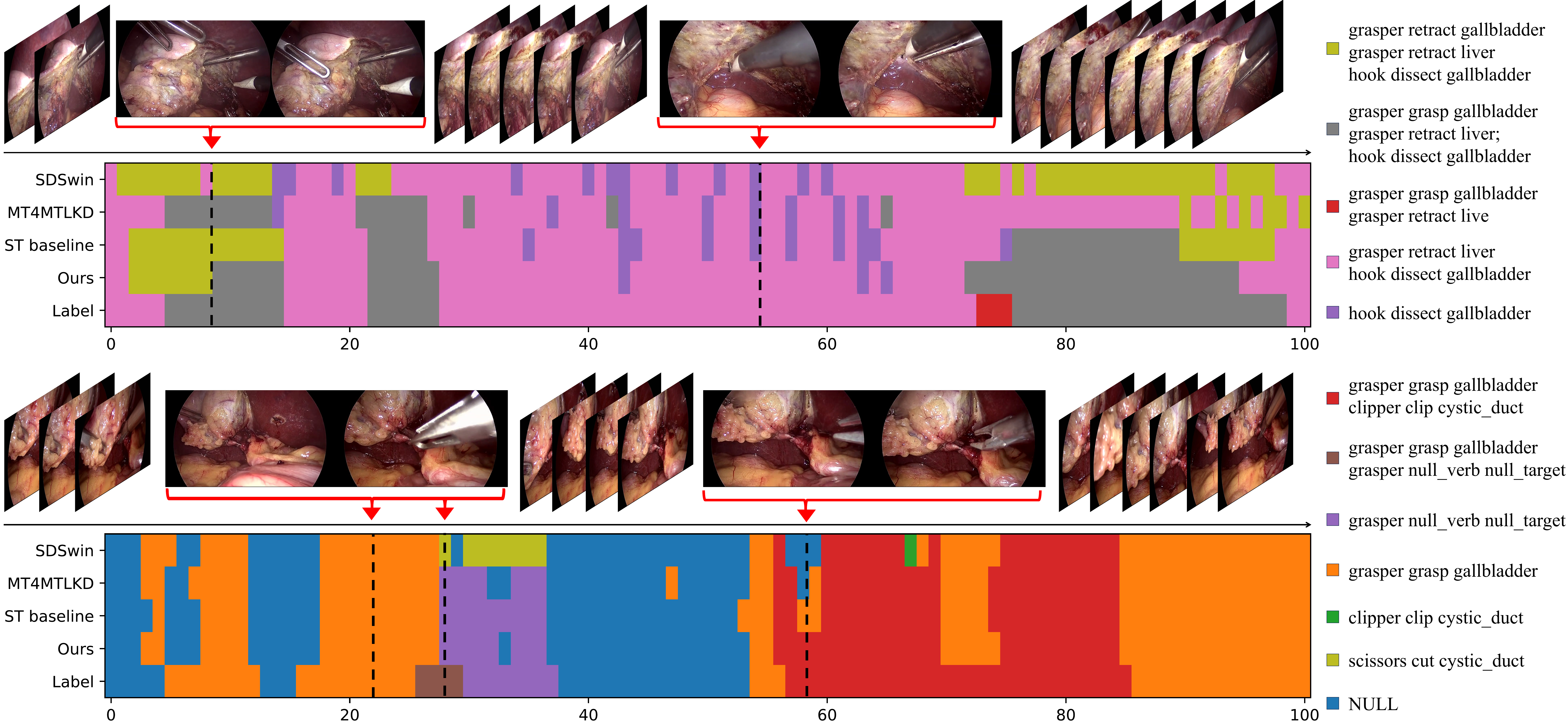}
    \caption{
    Qualitative visualization of surgical triplet recognition on a video sequence.
    The top row shows representative keyframes, while the bottom plot compares the frame-wise predictions of our method (Ours), the spatial–temporal baseline (ST baseline), and two state-of-the-art (SOTA) methods (SDSwin, MT4MTLKD) against the ground truth (Label). 
    Our approach produces triplet predictions that are semantically closer to the ground truth.
    }
    \vspace{-2mm}
    \label{images:Visualization}
\end{figure*}

\subsection{Compare with SOTA Methods}

We include state-of-the-art models specifically designed for surgical video understanding and triplet prediction, including RiT~\cite{sharma2023rendezvous}, RDV~\cite{nwoye2022rendezvous}, SDSwin~\cite{yamlahi2023self}, CoLSurgical~\cite{xi2023chain}, TERL~\cite{gui2024tail}, and MT4MTLKD~\cite{gui2023mt4mtl}.

\textbf{Quantitative Evaluation.} To validate the effectiveness of our method, we conduct quantitative evaluations on two public surgical video benchmarks and compare against recent state-of-the-art approaches (see \autoref{tab:Quantitative_AP_TCER}).

On the CholecT45 dataset, our framework achieves superior triplet recognition performance, with $AP_{IVT}$ increasing from 38.9\% to 40.9\%, corresponding to a relative improvement of 5.1\%. Meanwhile, our approach also leads in single-entity predictions. For Top-$k$ retrieval, the model performs competitively across different ranking cutoffs; it is only slightly surpassed by the best-performing baseline at Top-20. This consistency highlights the robustness of our triplet ranking mechanism. Notably, our model achieves the best results under the proposed TCER metrics, particularly in TCER$_{m}$, suggesting that the predicted IVT relations more faithfully preserve the semantic relationship among surgical entities.

On the multi-center ProstaTD dataset, our approach outperforms all baselines with 37.5\% $AP_{IVT}$ and lower TCER. Improved relational consistency further demonstrates its resilience to appearance variability.

\textbf{Qualitative Evaluation.} We provide a qualitative comparison on two representative video sequences in \autoref{images:Visualization}. For each sequence, the top row shows keyframes, while the bottom plot presents frame-wise predictions of our full method (Ours), the spatial-temporal baseline (ST baseline), and two SOTA methods (SDSwin, MT4MTLKD), alongside the ground truth (Label).

As illustrated in the highlighted segments of the video sequence, our model demonstrates improved robustness in challenging scenarios. 
In the second highlighted segment of the first sequence, where the instrument is partially occluded and visual cues are limited, competing approaches fail to maintain consistent triplet predictions, leading to rapid fluctuations and incorrect classifications. In contrast, our method preserves stable recognition results despite transient visual noise. We attribute this improvement to the introduction of semantic correlation modeling among triplets. By explicitly encoding inter-label dependencies, the model can infer plausible triplets even when visual evidence is incomplete or ambiguous. Similarly, in the first highlighted segment of the first sequence, our full model corrects an initial misclassification during an action transition, effectively leveraging learned label dependencies to refine temporal predictions. 
For the second sequence, in the third highlighted segments, the grasper is only partially visible in the upper-left corner of the frame; nonetheless, our method is the only one to successfully recognize it. In the first and second segments, however, all methods fail. Specifically, in the first segment, the triplet $<$grasper, grasp, gallbladder$>$ is present. As a new grasper enters the scene, the preceding one completely disappears from the field of view and remains invisible for several consecutive frames, leading to recognition failures across all methods. These observations indicate that purely temporal models are susceptible to occlusions or rapid motion in the absence of semantic constraints. By contrast, incorporating label relationships effectively mitigates this issue, yielding more stable and coherent triplet predictions.

We further provide qualitative comparisons against representative state-of-the-art methods, as shown in \autoref{fig:qualitative_triplet}. Each example displays the Top-5 predicted IVT triplets for a given surgical frame. Overall, our method demonstrates more accurate and reliable triplet recognition. 
In most cases, the proposed model successfully ranks the ground-truth surgical triplets within the Top-5 predictions, whereas competing approaches often fail to include all correct triplets in their Top-5 output. 
Moreover, even when all methods identify the correct triplets, our predictions typically achieve higher confidence ordering, reflecting stronger discriminative modeling of inter-component relationships. These qualitative findings are consistent with the quantitative improvements observed in $AP_{IVT}$ metrics. 
This leads to more clinically interpretable predictions.

\begin{figure*}[t]
    \centering
    \includegraphics[width=\textwidth]{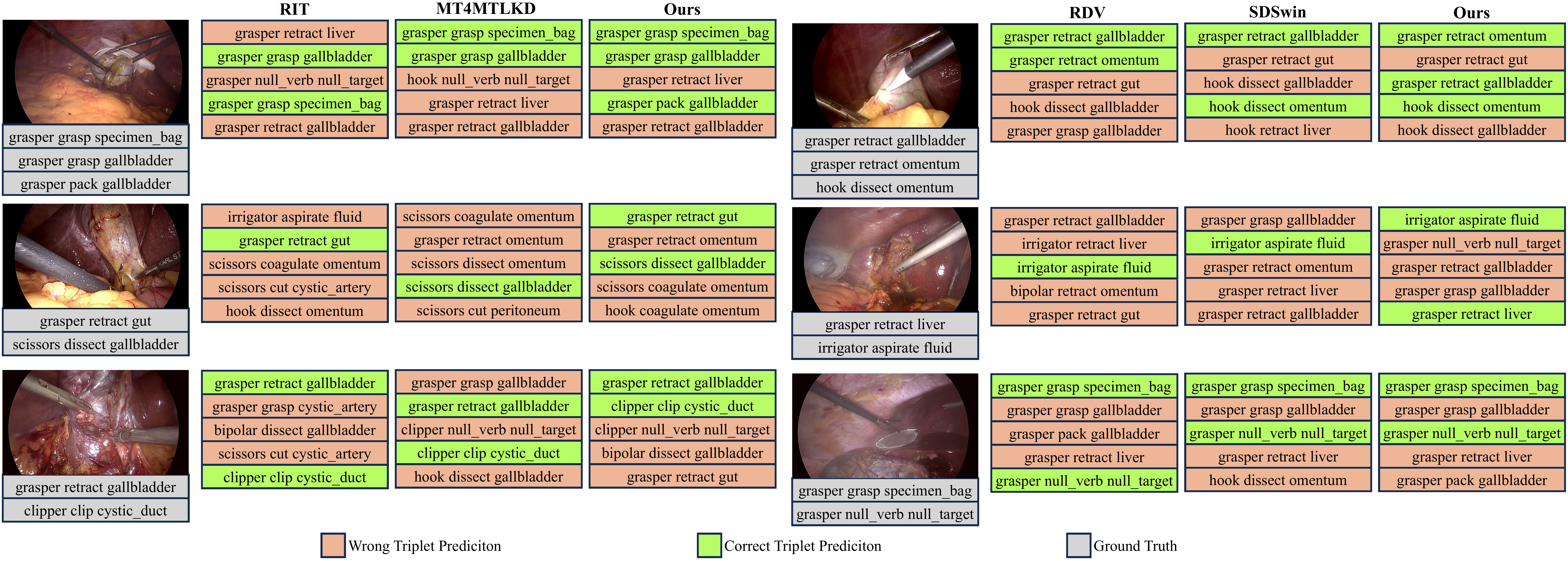}
    \caption{
        \textbf{Qualitative comparison of Top-5 triplet prediction results across different methods.}
        Each column corresponds to a representative model, including RIT~\cite{sharma2023rendezvous}, MT4MTLKD~\cite{gui2023mt4mtl}, RDV~\cite{nwoye2022rendezvous}, SDSwin~\cite{yamlahi2023self}, and our proposed framework.
        For each input surgical frame, the Top-5 predicted IVT triplets are shown in descending confidence order.
        Green boxes denote correctly predicted triplets, while red boxes indicate incorrect ones.
        Our method demonstrates superior semantic consistency and relational precision.}
    \label{fig:qualitative_triplet}
\end{figure*}

\subsection{Ablation Study}

\begin{table}[htbp]
\centering
\footnotesize
\renewcommand{\arraystretch}{1.25}
\setlength\tabcolsep{9pt}
\begin{tabular}{l|c|c|c}
\hline
\textbf{Config} & $AP_{IVT}$ $\uparrow$ & $TCER_{m}$ $\downarrow$ & $TCER_{c}$ $\downarrow$ \\
\hline
\multicolumn{4}{c}{\textit{Label Correlation Modeling (LCM) Ablation}} \\
\hline
W/o LCM & 37.9 & 6.65 & 2.67 \\
LCM Baseline  & 38.4 & 6.42 & 2.68 \\
\quad+ VAC & 39.2 & 5.70 & 2.25 \\
\quad+ MS-CAMRE & 38.7 & 6.24 & 2.33 \\
Full & \textbf{40.9} & \textbf{4.16} & \textbf{2.09} \\
\hline
\multicolumn{4}{c}{\textit{Fusion Strategy Ablation}} \\
\hline
Add & 39.5 & 5.32 & 2.44 \\
Concat & 39.7 & 5.37 & 2.32 \\
Single CA & 40.2 & 5.08 & 2.25 \\
BTRFA & \textbf{40.9} & \textbf{4.16} & \textbf{2.09} \\
\hline
\end{tabular}
\caption{
Ablation study on Label Correlation Modeling and Fusion strategies.
VAC enhances node semantic discrimination, MS-CAMRE strengthens relational reasoning,
while BTRFA enables more effective contextual interaction and representation fusion.
}
\label{table:all_ablation_revised}
\end{table}

\textbf{Ablation on Label Correlation Modeling.} As shown in \autoref{table:all_ablation_revised}, we conduct an ablation study to evaluate the contribution of each component in the LCM framework. The model without LCM lacks explicit label correlation modeling and therefore struggles to capture the interdependencies among triplets. Introducing static co-occurrence priors in the LCM Baseline provides only limited gains, as such priors cannot adapt dynamically to varying surgical contexts.

Based on the LCM Baseline, both VAC and MS-CAMRE independently enhance the model. Incorporating VAC improves performance by enhancing node-level semantic representations through CAMs, making individual label features more discriminative. MS-CAMRE refines the model by adaptively adjusting edge relations in the label graph, resulting in more context-consistent triplet predictions. When combined in the Full model, VAC and MS-CAMRE complement each other: VAC strengthens node semantics, while MS-CAMRE enforces proper relational structure across labels. This synergy achieves the highest $AP_{IVT}$ and the lowest TCER, highlighting the importance of jointly modeling node semantics and relational context.

As illustrated in \autoref{fig:edge}, the endoscopic frame (Left) contains co-occurrence between categories 7 (grasper grasp gallbladder) and 12 (grasper grasp specimen\_bag).

The LCM baseline (Middle), which relies solely on static priors, yields an adjacency matrix in which the interaction strength between the actually present categories is even lower than that between categories not present in the scene, indicating its inability to adapt to real-time visual context. In contrast, our MS-CAMRE-enhanced model (Right) effectively refines the GCN adjacency structure, substantially increasing the co-occurrence strength among the true relational categories while simultaneously suppressing spurious high-probability connections, thereby promoting a more reliable and context-consistent relational representation.

\begin{figure}[ht]
    \centering
    \includegraphics[width=0.48\textwidth]{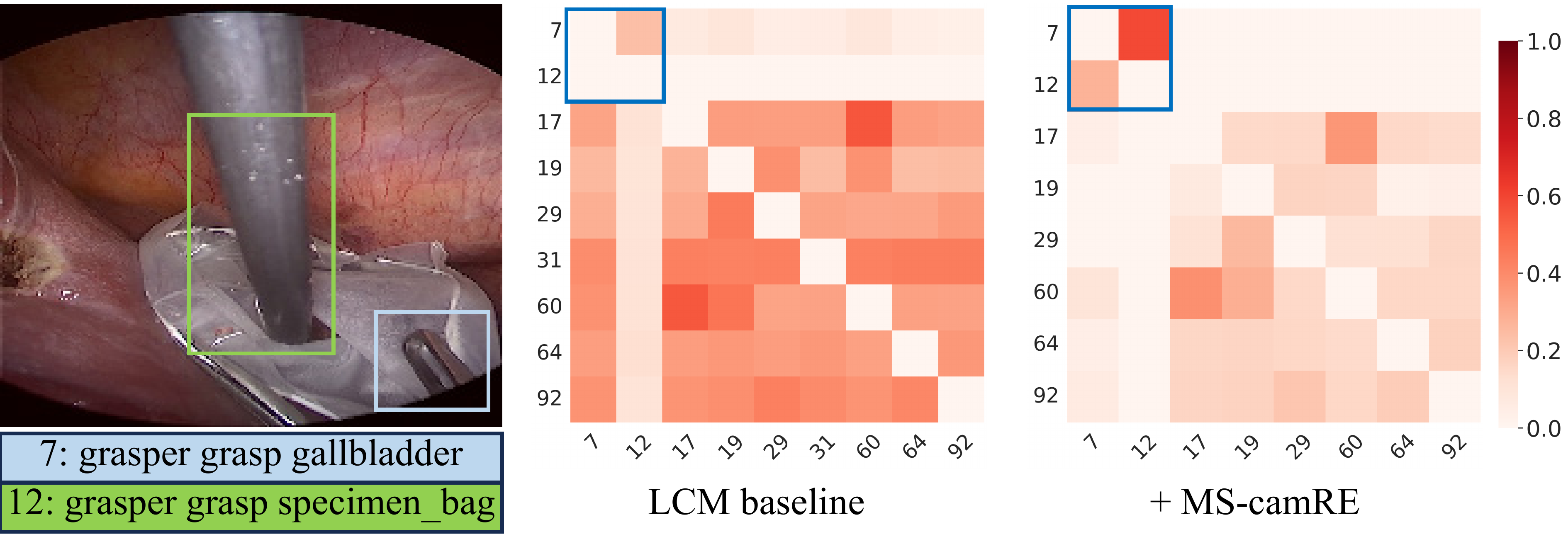}
    \caption{Comparison of MS-CAMRE and the LCM baseline. The proposed MS-CAMRE method adaptively refines the GCN adjacency matrix to more accurately boost the probabilities of truly co-occurring categories.}
    \label{fig:edge}
\end{figure}

\textbf{Ablation on BTRFA Module.} To evaluate the design of the BTRFA module, we compare it with three alternative fusion strategies: (1) Additive fusion, which directly sums bidirectional features; (2) Concatenation, which merges features before linear projection; and (3) Single Cross-Attention, which models temporal relations in only one direction. As shown in \autoref{table:all_ablation_revised}, while additive and concatenation strategies offer limited improvements, introducing cross-attention in a single direction yields further gains by selectively attending to temporally relevant cues. However, the full BTRFA achieves the best performance. This demonstrates the effectiveness of the proposed method.

\section{Conclusion and Limits}

In this work, we propose a unified framework for surgical triplet recognition that jointly models spatial, relational, and temporal cues.
The MS-CAMRE module captures both static and dynamic label dependencies and the BTRFA module fuses temporal and relational features for consistent multi-entity reasoning.
We also introduced a new metric, TCER, to evaluate relational consistency in surgical videos.
Extensive experiments on CholecT45 and ProstaTD demonstrate significant improvements in recognition accuracy. Despite these advances, our method has limitations: The structure and components of the model are too complex and training is difficult. Future work will explore adaptive memory updates and lightweight architectures for real-time surgical applications.

\label{sec:Conclusion}
\newpage
\begin{flushleft}
\textbf{Acknowledgement} This work was supported by the New Generation Artificial Intelligence-National Science and Technology Major Project (2025ZD0124000), the Beijing Natural Science Foundation under Grant (No. 4252018), the National Natural Science Foundation of China (No. 62572032), Beijing Natural Science Foundation (L242141) and the Shenzhen Science and Technology Program (Grant No. RCYX20231211090127030 and JCYJ20250604182948064).
\end{flushleft}

{
    \small
    \bibliographystyle{ieeenat_fullname}
    \bibliography{main}
}


\end{document}